\documentclass[10pt,twocolumn,letterpaper]{article}

\usepackage{iccv}
\usepackage{times}
\usepackage{epsfig}
\usepackage{graphicx}
\usepackage{amsmath}
\usepackage{amssymb}
\usepackage{multirow}
\usepackage{adjustbox}
\usepackage{mwe}
\usepackage{rotating}

\usepackage[pagebackref=true,breaklinks=true,letterpaper=true,colorlinks,bookmarks=false]{hyperref}

\iccvfinalcopy 

\def\iccvPaperID{67} 

\begin{document}

\title{Benchmarking and Error Diagnosis in Multi-Instance Pose Estimation}

\author{
Matteo Ruggero Ronchi \qquad \qquad \qquad \qquad \qquad Pietro Perona\\
{\tt\small \hspace{-5mm} \url{www.vision.caltech.edu/~mronchi}} \hspace{25mm} {\tt\small perona@caltech.edu}\\
California Institute of Technology, Pasadena, CA, USA
}

\maketitle
\thispagestyle{empty}

\vspace{-5mm}
\begin{abstract}
We propose a new method to analyze the impact of errors in algorithms for multi-instance pose estimation and a principled benchmark that can be used to compare them. We define and characterize three classes of errors - localization, scoring, and background - study how they are influenced by instance attributes and their impact on an algorithm's performance. Our technique is applied to compare the two leading methods for human pose estimation on the COCO Dataset, measure the sensitivity of pose estimation with respect to instance size, type and number of visible keypoints, clutter due to multiple instances, and the relative score of instances. The performance of algorithms, and the types of error they make, are highly dependent on all these variables, but mostly on the number of keypoints and the clutter. The analysis and software tools we propose offer a novel and insightful approach for understanding the behavior of pose estimation algorithms and an effective method for measuring their strengths and weaknesses.
\end{abstract}
\vspace{-5mm}

\section{Introduction}
\label{sec:intro}

Estimating the pose of a person from a single monocular frame is a challenging task due to many confounding factors such as perspective projection, the variability of lighting and clothing, self-occlusion, occlusion by objects, and the simultaneous presence of multiple interacting people. Nevertheless, the performance of human pose estimation algorithms has recently improved dramatically, thanks to the development of suitable deep architectures~\cite{bulat2016human,cao2016realtime,chen2014articulated, gkioxari2014using,newell2016stacked,papandreou2017towards,pishchulin2012articulated,Ramakrishna2014posemachines,wei2016convolutional,yang2011articulated} and the availability of well-annotated image datasets, such as \textit{MPII Human Pose Dataset} and \textit{COCO}~\cite{andriluka14cvpr,lin2014microsoft}. There is broad consensus that performance is saturated on simpler single-person datasets~\cite{Johnson10,Johnson11}, and researchers' focus is shifting towards less constrained and more challenging datasets~\cite{andriluka14cvpr,eichner2010we,lin2014microsoft}, where images may contain multiple instances of people, and a variable number of body parts (or keypoints) are visible.
\begin{figure}[t!]
\centering
\includegraphics[width=\linewidth]{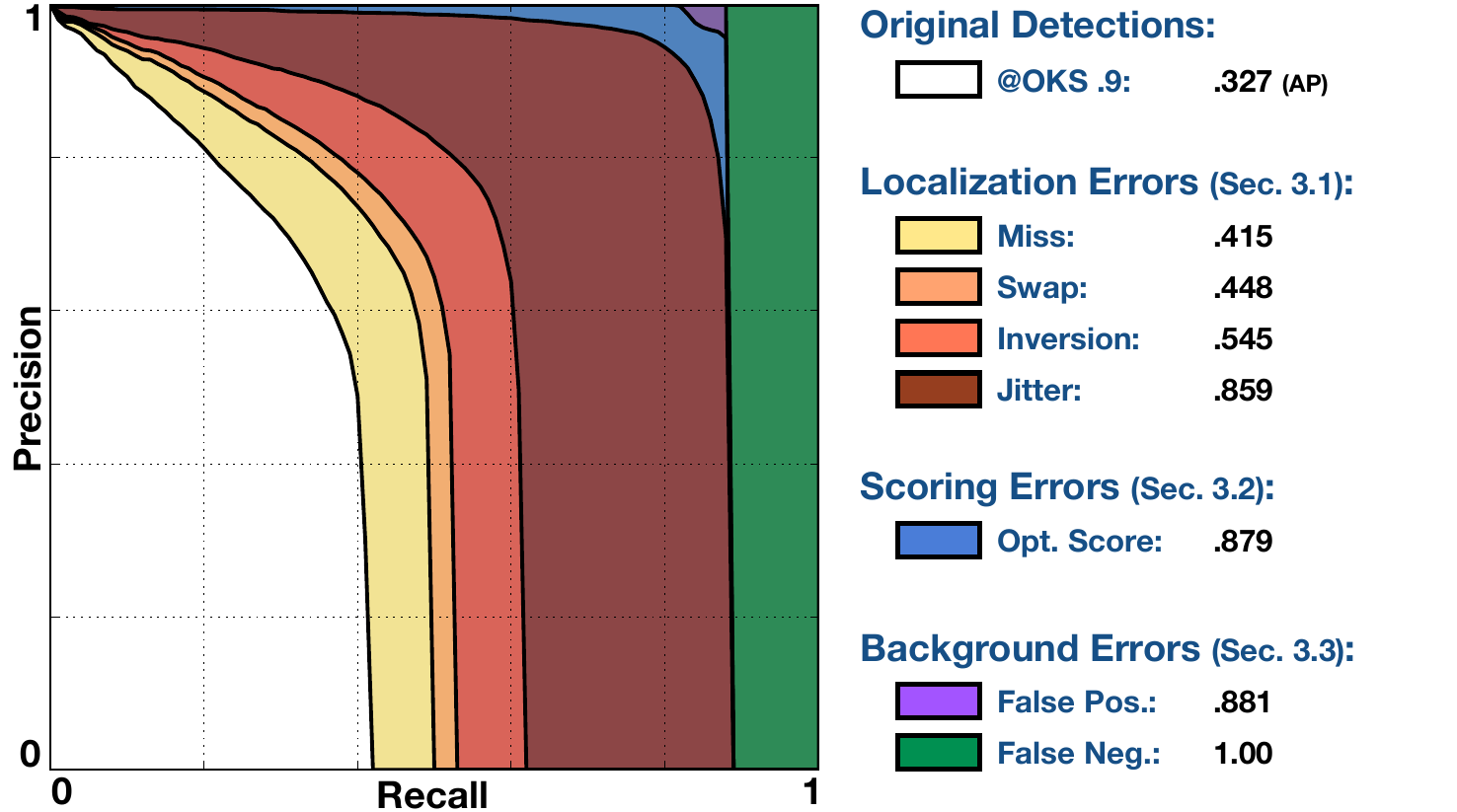}
{\caption {\small \textbf{Coarse to Fine Error Analysis.} We study the errors occurring in multi-instance pose estimation, and how they are affected by physical characteristics of the portrayed people. We build upon currently adopted evaluation metrics and provide the tools for a fine-grained description of performance, which allows to quantify the impact of different types of error at a single glance. The fine-grained Precision-Recall curves are obtained by fixing an OKS threshold and evaluating the performance of an algorithm after progressively correcting its mistakes.}}
\vspace{-5mm}
\label{fig:intro}
\end{figure}
However, evaluation is challenging: more complex datasets make it harder to benchmark algorithms due to the many sources of error that may affect performance, and existing metrics, such as Average Precision (AP) or mean Percentage of Correct Parts (mPCP), hide the underlying causes of error and are not sufficient for truly understanding the behaviour of algorithms.

Our goal is to propose a principled method for analyzing pose algorithms' performance. We make four contributions:\\
{\bf 1.} Taxonomization of the types of error that are typical of  the multi-instance pose estimation framework;\\
{\bf 2.} Sensitivity analysis of these errors with respect to measures of image complexity;\\
{\bf 3.} Side-by-side comparison of two leading human pose estimation algorithms highlighting key differences in behaviour that are hidden in the average performance numbers;\\
{\bf 4.} Assessment of which types of datasets and benchmarks may be most productive in guiding future research.

Our analysis extends beyond humans, to any object category where the location of parts is estimated along with detections, and to situations where cluttered scenes may contain multiple object instances. This is common in fine-grained categorization~\cite{branson2014bird}, or animal behavior analysis~\cite{CRIM13,eyjolfsdottir2014detecting}, where part alignment is often crucial.

\section{Related Work}
\label{sec:related_work}

\subsection{Error Diagnosis}
\label{subsec:error_analysis}

\textbf{Object Detection:} Hoiem et al.~\cite{hoiem2012diagnosing} studied how a detailed error analysis is essential for the progress of recognition research, since standard benchmark metrics do not tell us \textit{why} certain methods outperform others and \textit{how} could they be improved. They determined that several modes of failure are due to different types of error and highlighted the main confounding factors for object detection algorithms. While~\cite{hoiem2012diagnosing} pointed out the value of discriminating between different errors, it did not show how to do so in the context of pose estimation, which is one of our contributions.\vspace{1.5mm}

\textbf{Pose Estimation:} In their early work on pose regression, Doll{\'a}r et al.~\cite{dollar2010cascaded} observed that unlike human annotators, algorithms have a distribution of the normalized distances between a part detection and its ground-truth that is typically bimodal, highlighting the presence of multiple error modes. The \textit{MPII Human Pose Dataset}~\cite{andriluka14cvpr} Single-Person benchmark enables the evaluation of the performance of algorithms along a multitude of dimensions, such as 45 pose priors, 15 viewpoints and 20 human activities. However, none of the currently adopted benchmarks for Multi-Person pose estimation~\cite{eichner2010we,lin2014microsoft,pishchulin2016deepcut} carry out an extensive error and performance analysis specific to this framework, and mostly rely on the metrics from the Single-Person case. No standards for performing or compactly summarizing detailed evaluations has yet been defined, and as a result only a coarse comparison of algorithms can be carried out.

\subsection{Evaluation Framework}
\label{subsec:framework}

\begin{figure}[t!]
\centering
\includegraphics[width=\linewidth]{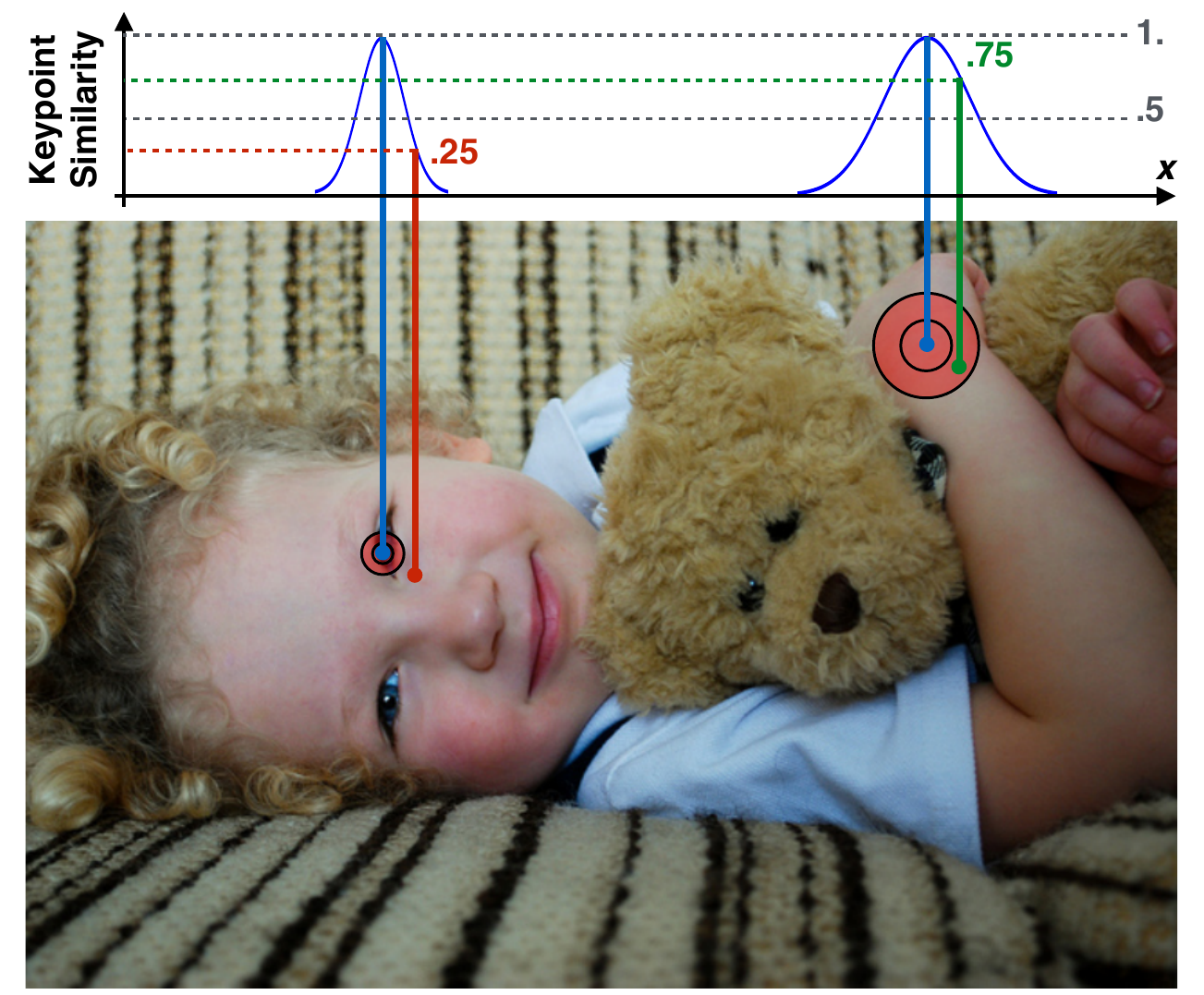}
\caption{ {\small \textbf{Keypoint Similarity (\textit{ks}).} The \textit{ks} between two detections, eye (red) and wrist (green), and their corresponding ground-truth (blue). The red concentric circles represent \textit{ks} values of .5 and .85 in the image plane and their size varies by keypoint type, see Sec.\ref{subsec:framework}. As a result, detections at the same distance from the corresponding ground-truth can have different \textit{ks} values.}}
\label{fig:keypoint_similarity}
\vspace{-11mm}
\end{figure}

We conduct our study on \textit{COCO}~\cite{lin2014microsoft} for several reasons: (i) it is the largest collection of multi-instance person keypoint annotations; (ii) performance on it is far from saturated and conclusions on such a large and non-iconic dataset can generalize to easier datasets; (iii) adopting their framework, with open source evaluation code, a multitude of datasets built on top of it, and annual competitions, will have the widest impact on the community.
The framework involves simultaneous person detection and keypoint estimation, and the evaluation mimics the one used for object detection, based on Average Precision and Recall (AP, AR). Given an image, a distance measure is used to match algorithm detections, sorted by their confidence score, to ground-truth annotations. For bounding-boxes and segmentations, the distance of a detection and annotation pair is measured by their Intersection over Union. In the keypoint estimation task, a new metric called Object Keypoint Similarity (OKS) is defined. The OKS between a detection $\mathbf{\hat{\theta}}^{(p)}$ and the annotation $\mathbf{\theta}^{(p)}$ of a person $p$, Eq.~\ref{eq:oks}, is the average over the labeled parts in the ground-truth ($v_i = 1,2$), of the \textit{Keypoint Similarity} between corresponding keypoint pairs, Fig.~\ref{fig:keypoint_similarity}; unlabeled parts ($v_i = 0$) do not affect the OKS~\cite{www_COCOEVAL}.

\begin{eqnarray}
\label{eq:oks}
\begin{cases}
ks(\mathbf{\hat{\theta}}_i^{(p)},\mathbf{\theta}_i^{(p)})  &= e^{-\frac{||\mathbf{\hat{\theta}}_i^{(p)}-\mathbf{\theta}_i^{(p)}||_{2}^2}{2s^2k_i^2}}\\
OKS(\mathbf{\hat{\theta}}^{(p)},\mathbf{\theta}^{(p)}) &= \frac{\sum_i ks(\mathbf{\hat{\theta}}_i^{(p)},\mathbf{\theta}_i^{(p)})\delta(v_i > 0)}{\sum_i\delta(v_i > 0)}
\end{cases}
\end{eqnarray}
\begin{figure*}[t!]
\centering
\begin{adjustbox}{max width=\linewidth}
\begin{tabular}[t]{cccccc}\hline\hline
\textbf{(Jitter)} & \multicolumn{2}{c}{\hspace{3mm}\textbf{(Inversion)}} & \multicolumn{2}{c}{\hspace{10mm}\textbf{(Swap)}} & \textbf{(Miss)}\\
\hline \vspace{-4mm}\\
\includegraphics[height=.3\linewidth]{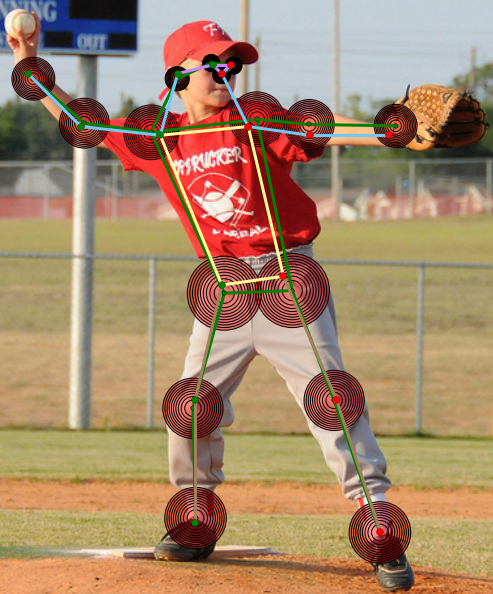} & 
\includegraphics[height=.3\linewidth]{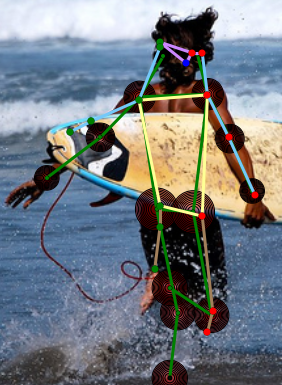} &
\includegraphics[height=.3\linewidth]{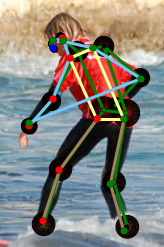} &
\includegraphics[height=.3\linewidth]{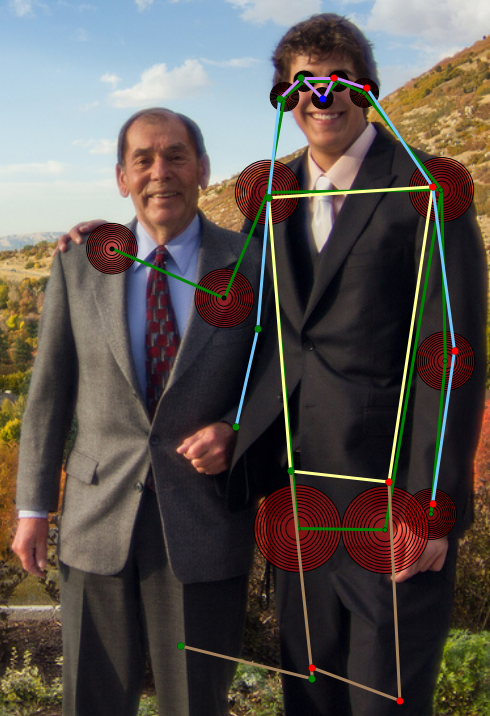} &
\includegraphics[height=.3\linewidth]{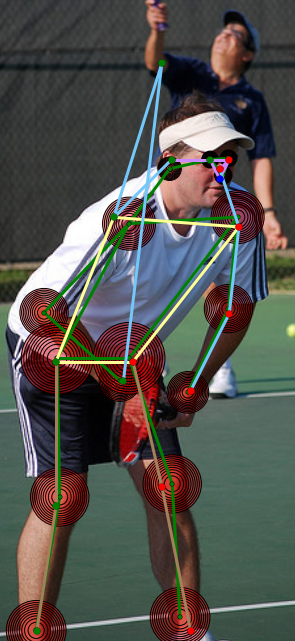} &
\includegraphics[height=.3\linewidth]{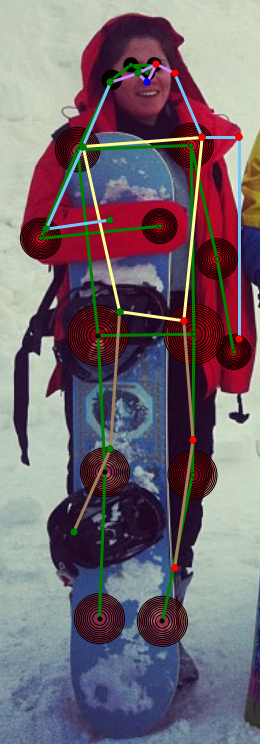} \\
(a) & (b) & (c) & (d) & (e) & (f)\\[1.25ex]
\end{tabular}
\end{adjustbox}
\caption{ {\small \textbf{Taxonomy of Keypoint Localization Errors.} Keypoint localization errors, Sec.~\ref{subsec:localization-errors}, are classified based on the position of a detection as, \textit{Jitter}: in the proximity of the correct ground-truth location, but not within the human error margin - left hip in (a); \textit{Inversion}: in the proximity of the ground-truth location of the wrong body part - inverted skeleton in (b), right wrist in (c); \textit{Swap}: in the proximity of the ground-truth location of the body part of a wrong person - right wrist in (d), right elbow in (e); \textit{Miss}: not in the proximity of any ground-truth location - both ankles in (f). While errors in (b,d) appear to be more excusable than those in (c,e) they have the same weight. Color-coding: (ground-truth) - concentric red circles centered on each keypoint's location connected by a green skeleton; (prediction) - red/green dots for left/right body part predictions connected with colored skeleton, refer to the Appendix for an extended description.}}
\label{fig:localization_errors}
\vspace{-5mm}
\end{figure*}
The \textit{ks} is computed by evaluating an un-normalized Gaussian function, centered on the ground-truth position of a keypoint, at the location of the detection to evaluate. The Gaussian's standard deviation $k_i$ is specific to the keypoint type and is scaled by the area of the instance $s$, measured in pixels, so that the OKS is a perceptually meaningful and easy to interpret similarity measure. For each keypoint type, $k_i$ reflects the consistency of human observers clicking on keypoints of type $i$ and is computed from a set of 5000 redundantly annotated images~\cite{www_COCOEVAL}. 

To evaluate an algorithm's performance, its detections within each image are ordered by confidence score and assigned to the ground-truths that they have the highest OKS with. As matches are determined, the pool of available annotations for lower scored detections is reduced. Once all matches have been found, they are evaluated at a certain OKS threshold (ranging from .5 to .95 in~\cite{www_ECCV16COCO}) and classified as True or False Positives (above or below threshold), and unmatched annotations are counted as False Negatives. Overall AP is computed as in the \textit{PASCAL VOC Challenge}~\cite{everingham2010pascal}, by sorting the detections across all the images by confidence score and averaging precision over a predefined set of 101 recall values. AR is defined as the maximum recall given a fixed number of detections per image~\cite{hosang2016makes}. Finally, we will refer to cocoAP and cocoAR when AP and AR are additionally averaged over all OKS threshold values (.5:.05:.95), as done in the \textit{COCO} framework~\cite{www_ECCV16COCO}.

\begin{figure}[t!]
\centering
\includegraphics[width=.95\linewidth]{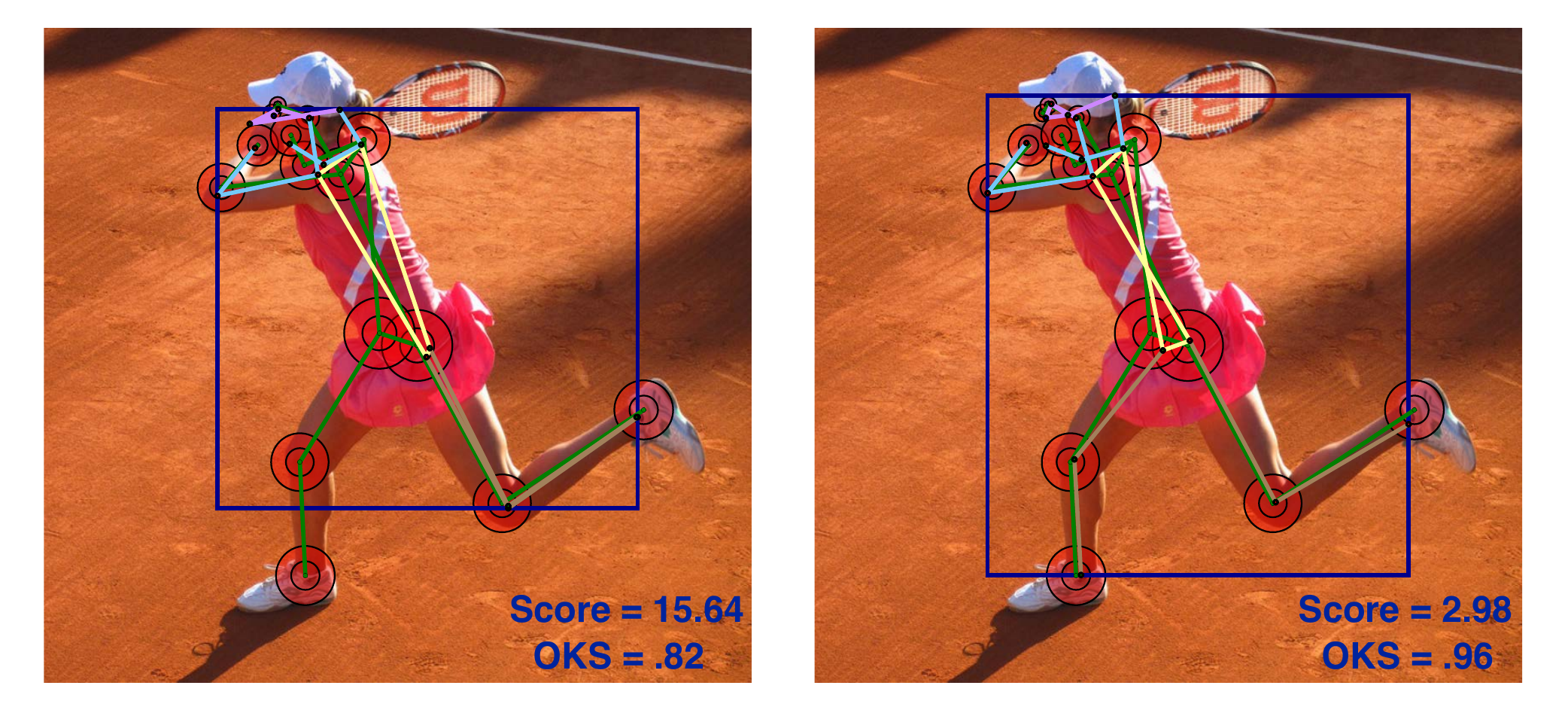}
\caption{ {\small \textbf{Instance Scoring Error.} The detection with highest confidence score (Left) is associated to the closest ground-truth by the evaluation algorithm described in Sec.~\ref{subsec:framework}. However, its OKS is lower than the OKS of another detection (Right). This results in a loss in performance at high OKS thresholds, details in Sec.~\ref{subsec:scoring_errors}.}}
\label{fig:scoring_errors}
\vspace{-5mm}
\end{figure}

\subsection{Algorithms}
\label{subsec:algorithms}

We conduct our analysis on the the top-two ranked algorithms~\cite{papandreou2017towards,cao2016realtime} of the \textit{2016 COCO Keypoints Challenge}~\cite{www_ECCV16COCO}, and observe the impact on performance of the design differences between a top-down and a bottom-up approach.

\textbf{Top-down (instance to parts)} methods first detect humans contained in an image, then try to estimate their pose separately within each bounding box~\cite{eichner2010we, gkioxari2014using, pishchulin2012articulated, yang2011articulated}. The \textbf{Grmi}~\cite{papandreou2017towards} algorithm is a two step cascade. In the first stage, a Faster-RCNN system~\cite{ren2015faster} using ResNet-Inception architecture~\cite{szegedy2016inception} combining inception layers~\cite{szegedy2015going} with residual connections~\cite{he2016deep} is used to produce a bounding box around each person instance. The second stage serves as a refinement where a ResNet with 101 layers~\cite{he2016deep} is applied to the image crop extracted around each detected person instance in order to localize its keypoints. The authors adopt a combined classification and regression approach~\cite{szegedy2014scalable,ren2015faster}: for each spatial position, first a classification problem is solved to determine whether it is in the vicinity of each of the keypoints of the human body, followed by a regression problem to predict a local offset vector for a more precise estimate of the exact location. The results of both stages are aggregated to produce highly localized activation maps for each keypoint in the form of a voting process: each point in a detected bounding box casts a vote with its estimate for the position of every keypoint, and the vote is weighted by the probability that it lays near the corresponding keypoint.

\begin{table}[t]
\setlength{\tabcolsep}{4pt}
\caption{\textit{2016 COCO Keypoints Challenge Leaderboard}~\cite{www_ECCV16COCO}}
\medskip
\label{tab:eccv_results}
\begin{adjustbox}{max width=\linewidth}
\begin{tabular}{lccccc}
\hline\noalign{\smallskip}
 & \textbf{Cmu} & Grmi & DL61 & R4D & Umichvl\\
\noalign{\smallskip}
\hline
\noalign{\smallskip}
cocoAP \quad\quad\quad & 0.608 & 0.598 & 0.533 & 0.497 & 0.434\\
\hline
\end{tabular}
\end{adjustbox}
\setlength{\tabcolsep}{1.4pt}
\vspace{-7mm}
\end{table}

\textbf{Bottom-up (parts to instance)} methods first separately detect all the parts of the human body from an image, then try to group them into individual instances~\cite{bulat2016human, newell2016stacked, Ramakrishna2014posemachines, wei2016convolutional}. The \textbf{Cmu}~\cite{cao2016realtime} algorithm estimates the pose for all the people in an image by solving body part detection and part association jointly in one end-to-end trainable network, as opposed to previous approaches that train these two tasks separately~\cite{insafutdinov2016deepercut, pishchulin2016deepcut} (typically part detection is followed by graphical models for the association). Confidence maps with gaussian peaks in the predicted locations, are used to represent the position of individual body parts in an image. Part Affinity Fields (PAFs) are defined from the confidence maps, as a set of 2D vector fields that jointly encode the \textit{location} and \textit{orientation} of a particular limb at each position in the image. The authors designed a two-branch VGG~\cite{simonyan2014very} based architecture, inspired from CPMs~\cite{wei2016convolutional}, to iteratively refine confidence maps and PAFs with global spatial contexts. The final step consists of a maximum weight bipartite graph matching problem~\cite{west2001introduction,kuhn1955hungarian} to associate body parts candidates and assemble them into full body poses for all the people in the image. A greedy association algorithm over a minimum spanning tree is used to group the predicted parts into consistent instance detections.


\begin{figure*}[t!]
\centering
\includegraphics[width=\linewidth]{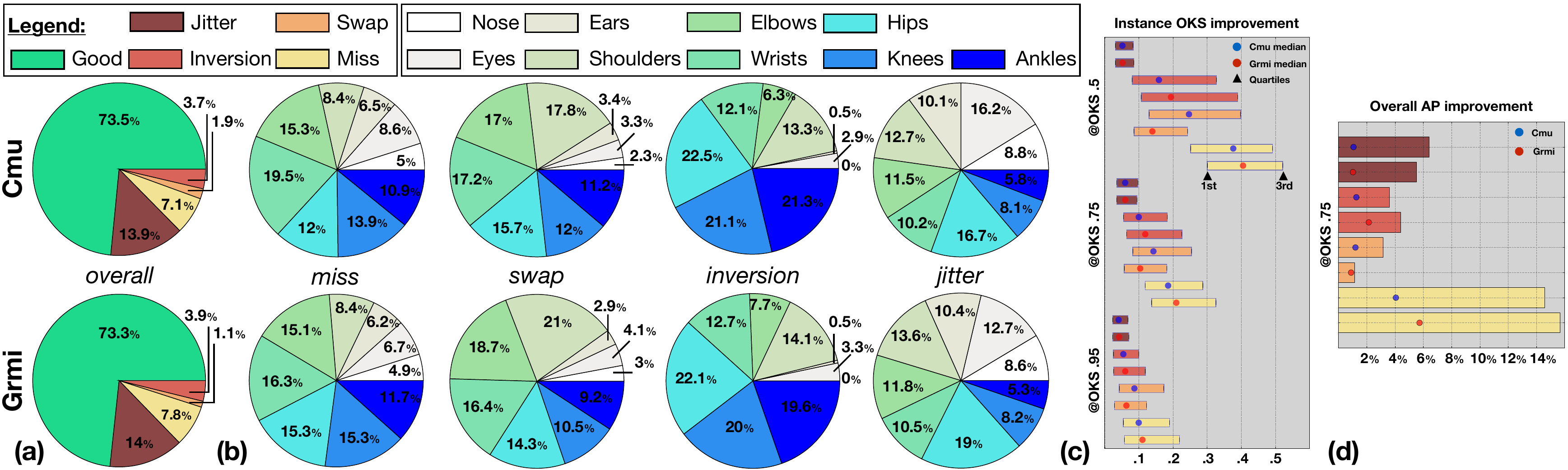}
\caption{ {\small \textbf{Distribution and Impact of Localization Errors.} (a) Outcome for the predicted keypoints: \textit{Good} indicates correct localization. (b) The breakdown of errors over body parts. (c) The algorithm's detections OKS improvement obtained after separately correcting errors of each type; evaluated over all the instances at OKS thresholds of .5, .75 and .95; the dots show the median, and the bar limits show the first and third quartile of the distribution. (d) The AP improvement obtained after correcting localization errors; evaluated at OKS thresholds of .75 (bars) and .5 (dots). A larger improvement in (c) and (d) shows what errors are more impactful. See Sec.~\ref{subsec:localization-errors} for details.}}
\label{fig:localization_errors_impact}
\vspace{-5mm}
\end{figure*}

\section{Multi-Instance Pose Estimation Errors}
\label{sec:analysis}

We propose a taxonomy of errors specific to the multi-instance pose estimation framework: (i) \textbf{Localization}, Fig.~\ref{fig:localization_errors}, due to the poor localization of the keypoint predictions belonging to a detected instance; (ii) \textbf{Scoring}, Fig.~\ref{fig:scoring_errors}, due to a sub-optimal confidence score assignment; (iii) \textbf{Background False Positives}, detections without a ground-truth annotation match; (iv) \textbf{False Negatives}, missed detections. We assess the causes and impact on the behaviour and performance of~\cite{cao2016realtime,papandreou2017towards} for each error type.

\subsection{Localization Errors}
\label{subsec:localization-errors}

A localization FP occurs when the location of the keypoints in a detection results in an OKS score with the corresponding ground-truth match that is lower than the evaluation threshold. They are typically due to the fact that body parts are difficult to detect because of self occlusion or occlusion by other objects. We define four types of localization errors, visible in Fig.~\ref{fig:localization_errors}, as a function of the keypoint similarity $ks(.,.)$, Eq.~\ref{eq:oks}, between the keypoint $i$ of a detection $\mathbf{\hat{\theta}}_i^{(p)}$ and $j$ of the annotation $\mathbf{\theta}_j^{(p)}$ of a person $p$.\\

\vspace{-2mm}
\textbf{Jitter:} small error around the correct keypoint location.
\vspace{-2mm}
$$.5 \leq ks(\mathbf{\hat{\theta}}_i^{(p)},\mathbf{{\theta}}_i^{(p)}) < .85$$
The limits can be chosen based on the application of interest; in the \textit{COCO} framework, .5 is the smallest evaluation threshold, and .85 is the threshold above which also human annotators have a significant disagreement (around $30\%$) in estimating the correct position~\cite{www_COCOEVAL}.\\

\textbf{Miss:} large localization error, the detected keypoint is not within the proximity of any body part.
$$ks(\mathbf{\hat{\theta}}_i^{(p)},\mathbf{{\theta}}_j^{(q)}) < .5 \quad \forall {q} \in \mathcal{P} \quad \textrm{and} \quad \forall {j} \in \mathcal{J}$$

\textbf{Inversion:} confusion between semantically similar parts belonging to the same instance. The detection is in the proximity of the true keypoint location of the wrong body part.
\begin{align*}
    ks(\mathbf{\hat{\theta}}_i^{(p)},\mathbf{{\theta}}_i^{(p)}) < .5\\
    \exists {j} \in \mathcal{J} \quad | \quad ks(\mathbf{\hat{\theta}}_i^{(p)},\mathbf{{\theta}}_{j}^{(p)}) \geq .5
\end{align*}

In our study we only consider inversions between the left and right parts of the body, however, the set of keypoints $\mathcal{J}$ can be arbitrarily defined to study any kind of inversion.

\textbf{Swap:} confusion between semantically similar parts of different instances. The detection is within the proximity of a body part belonging to a different person.
\vspace{-2mm}
\begin{align*}
  ks(\mathbf{\hat{\theta}}_i^{(p)},\mathbf{{\theta}}_i^{(p)}) < .5\\
    \exists {j} \in \mathcal{J} \quad \textrm{and} \quad \exists {q} \in \mathcal{P} \quad | \quad ks(\mathbf{\hat{\theta}}_i^{(p)},\mathbf{{\theta}}_{j}^{(q)}) \geq .5
\end{align*}

Every keypoint detection having a keypoint similarity with its ground-truth that exceeds .85 is considered \textit{good}, as it is within the error margin of human annotators. We can see, Fig.~\ref{fig:localization_errors_impact}.(a), that about $75\%$ of both algorithm's detections are \textit{good}, and while the percentage of \textit{jitter} and \textit{inversion} errors is approximately equal, \cite{cao2016realtime} has twice as many \textit{swaps}, and \cite{papandreou2017towards} has about $1\%$ more \textit{misses}. Fig.~\ref{fig:localization_errors_impact}.(b) contains a breakdown of errors over keypoint type: faces are easily detectable (smallest percentage of \textit{miss} errors); \textit{swap} errors are focused on the upper-body, as interactions typically involve some amount of upper-body occlusion; the lower-body is prone to \textit{inversions}, as people often self-occlude their legs, and there are less visual cues to distinguish left from right; finally \textit{jitter} errors are predominant on the hips. There are no major differences between the two algorithms in the above trends, indicating that none of the methods contains biases over keypoint type. After defining and identifying localization errors, we measure the improvement in performance resulting from their correction.

Localization errors are corrected by repositioning a keypoint prediction at a distance from the true keypoint location equivalent to a \textit{ks} of .85 for \textit{jitter}, .5 for \textit{miss}, and at a distance from the true keypoint location equivalent to the prediction's distance from the wrong body part detected in the case of \textit{inversion} and \textit{swap}\footnote{The Appendix contains examples showing how errors are corrected.}. Correcting localization errors results in an improvement of the OKS of every instance and the overall AP, as some detections become True Positives (TP) because the increased OKS value exceeds the evaluation threshold. Fig.~\ref{fig:localization_errors_impact}.(c) shows the OKS improvement obtainable by correcting errors of each type: it is most important to correct \textit{miss} errors, followed by \textit{inversions} and \textit{swaps}, while \textit{jitter} errors, although occurring most frequently, have a small impact on the OKS. We learn, Fig.~\ref{fig:localization_errors_impact}.(d), that \textit{misses} are the most costly error in terms of AP ($\sim15\%$), followed by \textit{inversions} ($\sim4\%$), relative to their low frequency. We focus on the improvement at the .75 OKS threshold, as it has almost perfect correlation with the value of cocoAP (average of AP over all thresholds)~\cite{huang2016speed}. Changing the evaluation threshold changes the impact of errors (for instance by lowering it to .5 more detections are TP so there is less AP improvement from their correction), but the same relative trends are verified, indicating that the above observations reflect the behavior of the methods and are not determined by the strictness of evaluation.

\subsection{Scoring Errors}
\label{subsec:scoring_errors}
\begin{figure}[t!]
\centering
\includegraphics[width=\linewidth]{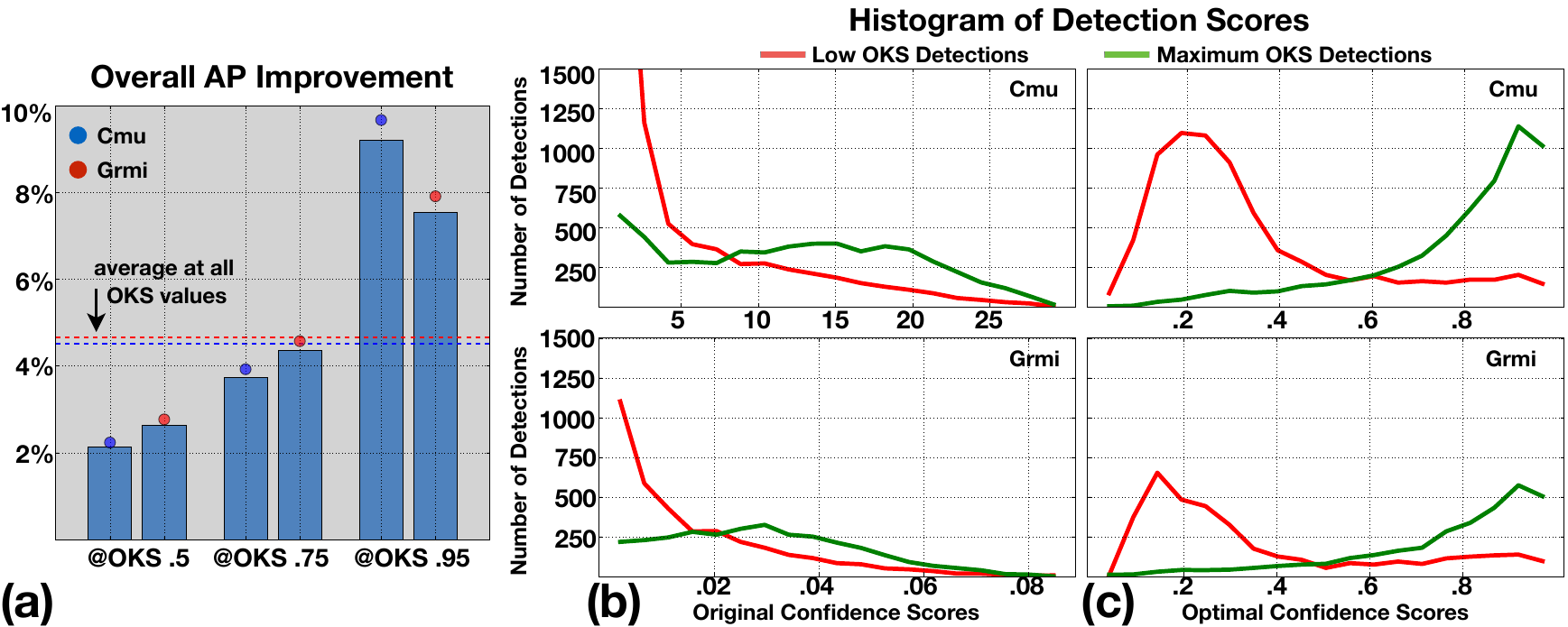}
\caption{ {\small \textbf{Scoring Errors Analysis.} (a) The AP improvement obtained when using the optimal detection scores, as defined in Sec.~\ref{subsec:scoring_errors}. The histogram of detections' (b) original and (c) optimal confidence scores. We histogram separately the scores of detections achieving the maximum OKS with a given ground-truth instance (green) and the other detections achieving OKS of at least .1 (red). High overlap of the histograms, as in (b), is caused by the presence of many detections with high OKS and low score or vice versa; a large separation, as in (c), is indication of a better score.}}
\label{fig:scoring_errors_impact}
\vspace{-5mm}
\end{figure}

Assigning scores to instances is a typical task in object detection, but a novel challenge for keypoint estimation. 
A scoring error occurs when two detections $\mathbf{\hat{\theta}}_{1}^{(p)}$ and $\mathbf{\hat{\theta}}_{2}^{(p)}$ are in the proximity of a ground-truth annotation $\mathbf{{\theta}}^{(p)}$ and the one with the highest confidence has the lowest OKS:
\vspace{-1mm}
$$\begin{cases}
   Score(\mathbf{\hat{\theta}}_{1}^{(p)}) &> Score(\mathbf{\hat{\theta}}_{2}^{(p)})\\
   OKS(\mathbf{\hat{\theta}}_{1}^{(p)},\mathbf{{\theta}}^{(p)}) &< OKS(\mathbf{\hat{\theta}}_{2}^{(p)},\mathbf{{\theta}}^{(p)})
\end{cases}$$

\begin{table}[t]
\setlength{\tabcolsep}{4pt}
\caption{Improvements due to the optimal rescoring of detections.}
\label{tab:scoring_errors_impact}
\begin{adjustbox}{max width=\linewidth}
\begin{tabular}{lcc}
\hline\noalign{\smallskip}
\textbf{} & \textbf{Cmu}~\cite{cao2016realtime} & \textbf{Grmi}~\cite{papandreou2017towards}\\
\noalign{\smallskip}
\hline
\noalign{\smallskip}
Imgs. w. detections        & 11940         & 14634 \\
Imgs. w. optimal detection order     & 7456 (62.4\%) & 9934 (67.8\%)\\
Number of Scoring Errors     & 407           & 82\\
Increase of Matches          & 64            & 156\\
Matches with OKS Improvement & 590           & 430\\
\hline
\end{tabular}
\end{adjustbox}
\setlength{\tabcolsep}{1.4pt}
\vspace{-7mm}
\end{table}
This can happen in cluttered scenes when many people and their detections are overlapping, or in the case of an isolated person for which multiple detections are fired, Fig.~\ref{fig:scoring_errors}. Confidence scores affect evaluation, Sec.~\ref{subsec:framework}, \textit{locally} by determining the order in which detections get matched to the annotations in an image, and \textit{globally}, when detections are sorted across the whole dataset to compute AP and AR. As a result, it is important for the detection scores to be: (i) \textit{`OKS monotonic increasing'}, so that a higher score always results in a higher OKS; (ii) calibrated, so that scores reflect as much as possible the probability of being a TP. A score possessing such properties is \textit{optimal}, as it achieves the highest performance possible for the provided detections. It follows that the optimal score for a given detection corresponds to the maximum OKS value obtainable with any ground-truth annotation: monotonicity and perfect calibration are both guaranteed, as higher OKS detections would have higher score, and the OKS is an exact predictor of the quality of a detection. The optimal scores can be computed at evaluation time, by an oracle assigning to each detection a confidence corresponding to the maximum OKS score achievable with any ground-truth instance. To aid performance in the case of strong occlusion, we apply Soft-Non-Max-Suppression~\cite{bodla2017improving}, which decays the confidence scores of detections as a function of the amount of reciprocal overlap.

Using optimal scores yields about $5\%$ AP improvement, averaged at all the OKS evaluation thresholds, and up to $10\%$ at OKS .95, Fig.~\ref{fig:scoring_errors_impact}.(a), pointing to the importance of assigning low scores to unmatched detections. A careful examination shows that the reason of the improvement is two-fold, Tab.~\ref{tab:scoring_errors_impact}: (i) there is an increase in the number of matches between detections and ground-truth instances (reduction of \textit{FP} and \textit{FN}) and (ii) the existing matches obtain a higher OKS value. Both methods have a significant amount of overlap, Fig.~\ref{fig:scoring_errors_impact}.(b), between the histogram of \textit{original} scores for the detections with the highest OKS with a given ground-truth (green line) and all other detections with a lower OKS (red line). This indicates the presence of many detections with high OKS and low score or vice versa. Fig.~\ref{fig:scoring_errors_impact}.(c) shows the effect of rescoring: \textit{optimal score} distributions are bi-modal and present a large separation, so confidence score is a better OKS predictor. Although the AP improvement after rescoring is equivalent, \cite{papandreou2017towards} provides scores that are in the same order as the optimal ones for a higher percentage of images and makes less errors, indicating that it is using a better scoring function. 
\begin{figure*}[!t]
\centering
\includegraphics[width=\linewidth]{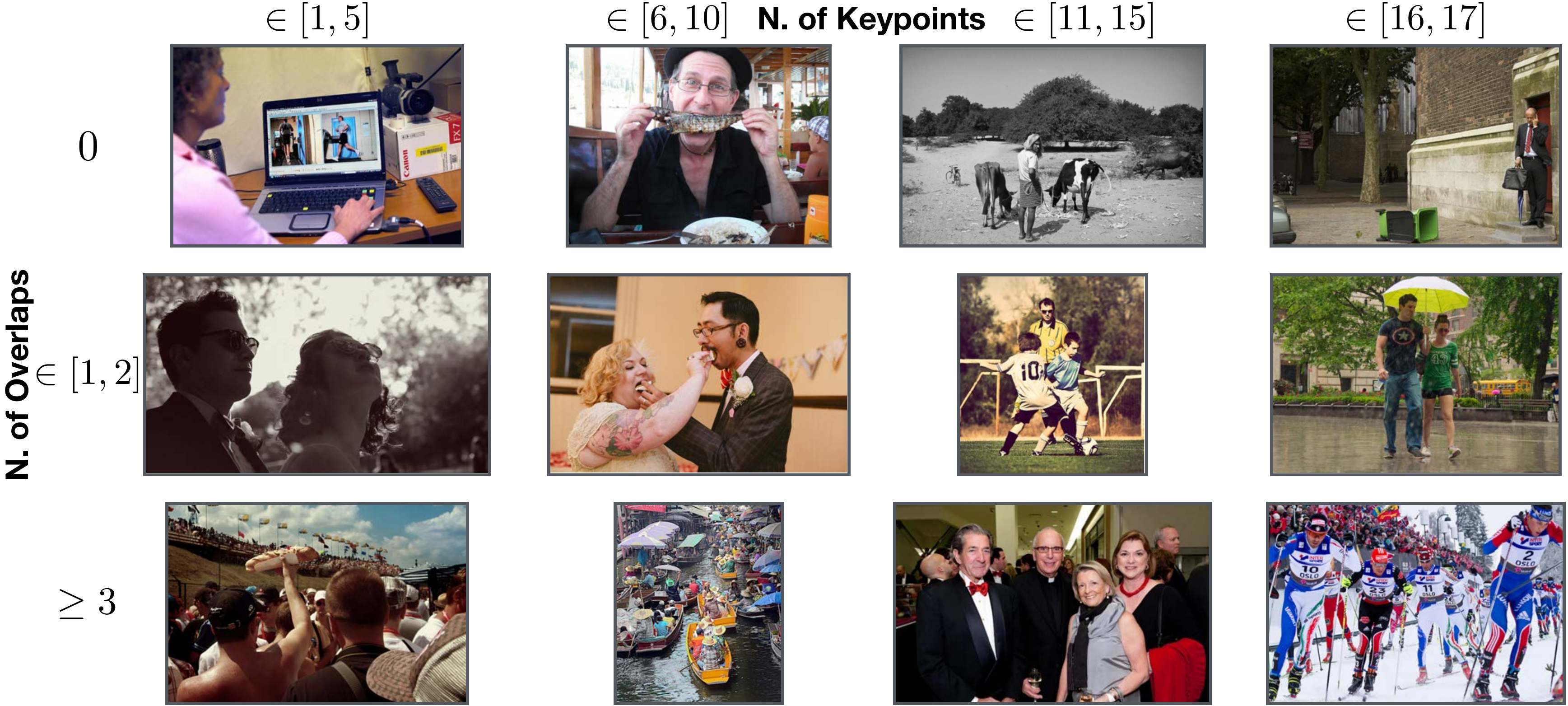}\\[.15ex]
\caption{ {\small \textbf{Images from the \textit{COCO} Dataset Benchmarks.} We separate the ground-truth instances in the \textit{COCO} dataset into twelve benchmarks, based on number of visible keypoints and overlap between annotations; Fig.~\ref{fig:coco_benchmark_stats}.(b) shows the size of each benchmark.}}
\label{fig:coco_benchmark}
\vspace{-5mm}
\end{figure*}
\subsection{Background False Positives and False Negatives}
\label{sec:background_errors}

\begin{figure}[t!]
\centering
\includegraphics[width=\linewidth]{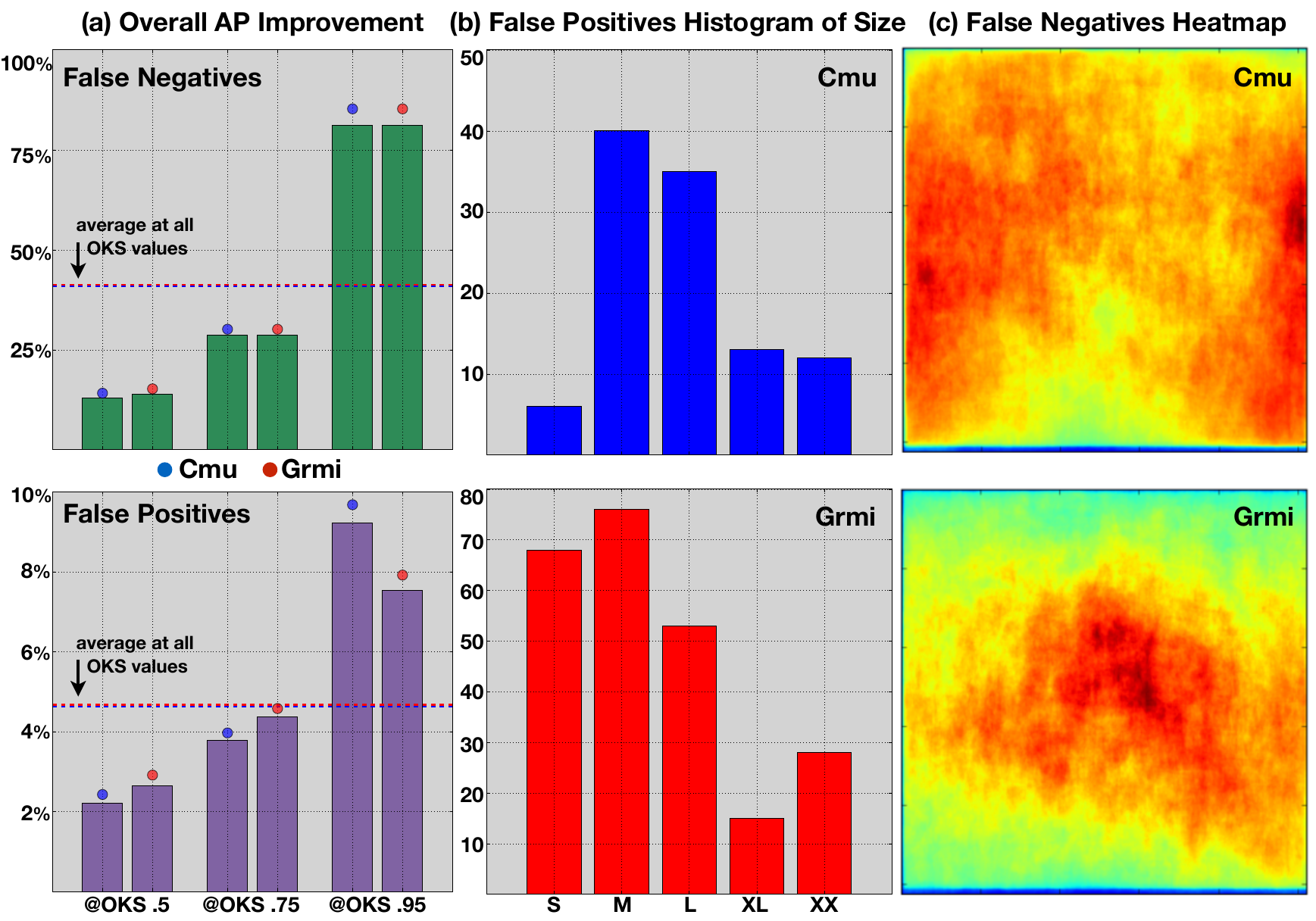}
\caption{ {\small \textbf{Background Errors Analysis.} (a) The AP improvement obtained after \textit{FN} (top) and \textit{FP} (bottom) errors are removed from evaluation; horizontal lines show the average value for each method. (b) The histogram of the area size of \textit{FP} having a high confidence score. (c) The heatmaps obtained by adding the resized ground-truth \textit{COCO} segmentation masks of all the \textit{FN}.}}
\label{fig:background_false_negatives}
\vspace{-5mm}
\end{figure}

\textit{FP} and \textit{FN} respectively consist of an algorithm's detections and the ground-truth annotations that remain unmatched after evaluation is performed. \textit{FP} typically occur when objects resemble human features or when body parts of nearby people are merged into a wrong detection. Most of the \textit{FP} errors could be resolved by performing better Non-Max-Suppression and scoring, since their impact is greatly reduced when using optimal scores, i.e. Fig.~\ref{fig:intro}. Small size and low number of visible keypoints are instead the main cause of \textit{FN}. In Fig.~\ref{fig:background_false_negatives}.(a) we show the impact of background errors on the AP at three OKS evaluation thresholds: FN affect performance significantly more than \textit{FP}, on average about $40\%$ versus only $5\%$. For both methods, the average number of people in images containing \textit{FP} and \textit{FN} is about $5$ and $7$, compared to the dataset's average of $3$, suggesting that cluttered scenes are more prone to having background errors. Interestingly, the location of \textit{FN} errors for the two methods differs greatly, Fig.\ref{fig:background_false_negatives}.(c):~\cite{cao2016realtime} predominantly misses annotations around the image border, while~\cite{papandreou2017towards} misses those at the center of an image. Another significant difference is in the quantity of \textit{FP} detections having a high confidence score (in the top-20th percentile of overall scores), Fig.\ref{fig:background_false_negatives}.(b):~\cite{papandreou2017towards} has more than twice the number, mostly all with small pixel area size ($<32^{2}$).

\section{Sensitivity to Occlusion, Crowding and Size}
\label{sec:coco_benchmark}
\begin{figure*}[t!]
\centering
\resizebox{\linewidth}{!}{
\begin{tabular}{cc}
\includegraphics[height=.7\linewidth]{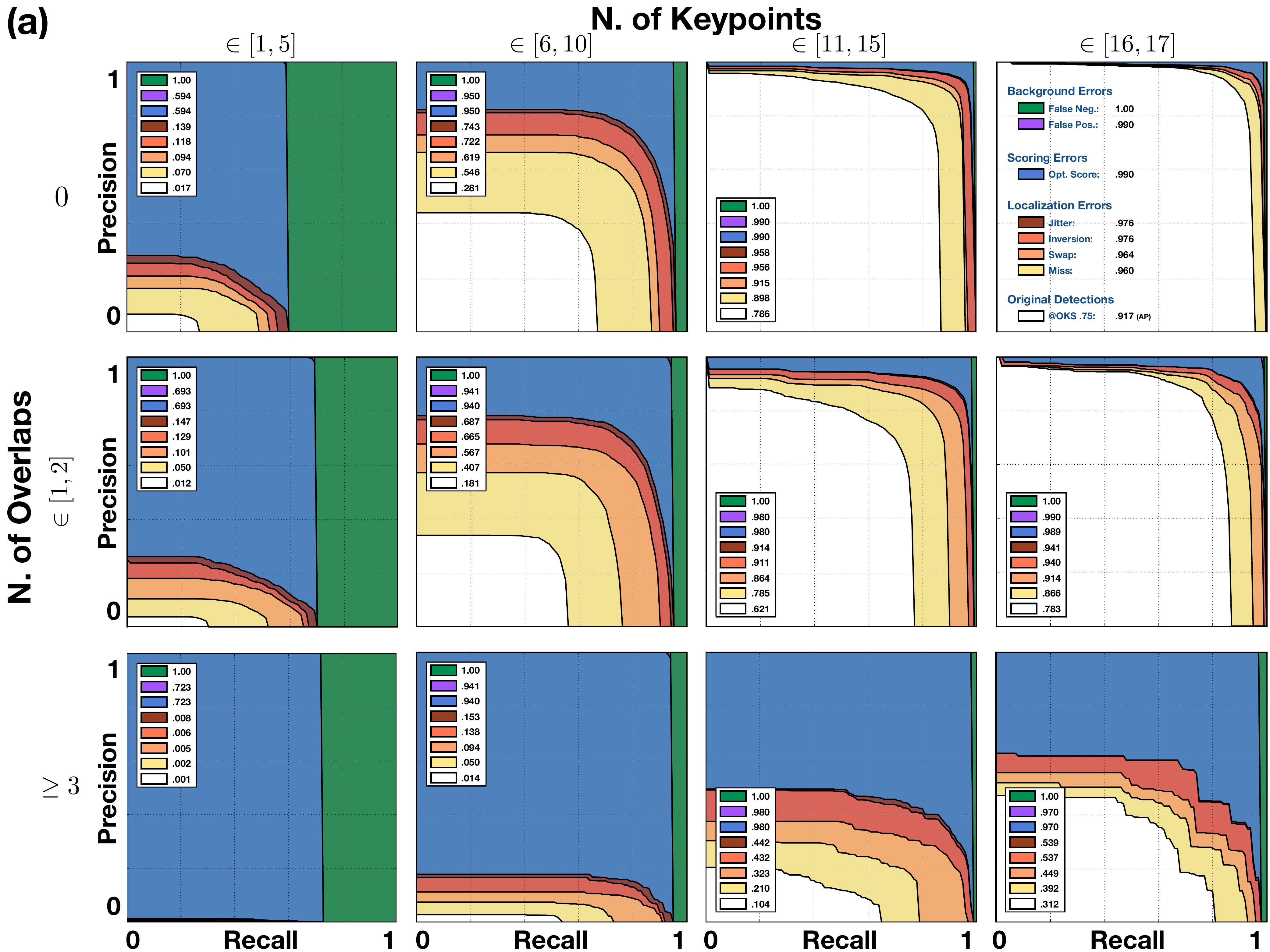} & 
\raisebox{2mm}{\includegraphics[height=.68\linewidth]{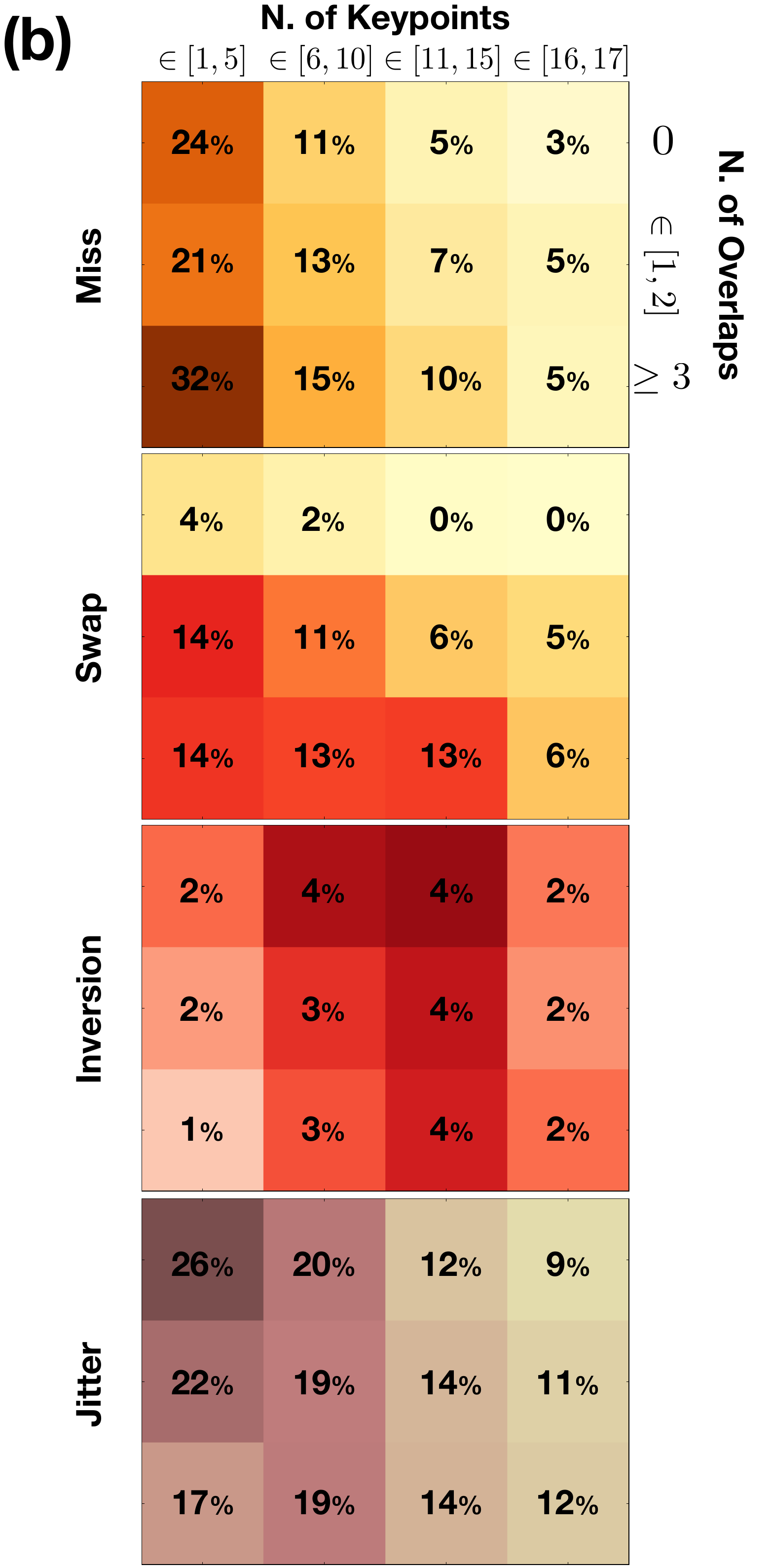}}\\[.15ex]
\end{tabular}
}
\caption{ {\small \textbf{Performance and Error Sensitivity to Occlusion and Crowding.} (a) The PR curves showing the performance of~\cite{cao2016realtime} obtained by progressively correcting errors of each type at the OKS evaluation threshold of .75 on the twelve Occlusion and Crowding Benchmarks described in Sec.~\ref{sec:coco_benchmark}; every legend contains the overall AP values. (b) The frequency of localization errors occurring on each benchmark set.}}
\label{fig:coco_benchmark_impact}
\vspace{-5mm}
\end{figure*}
\begin{figure}[t!]
\centering
\includegraphics[width=\linewidth]{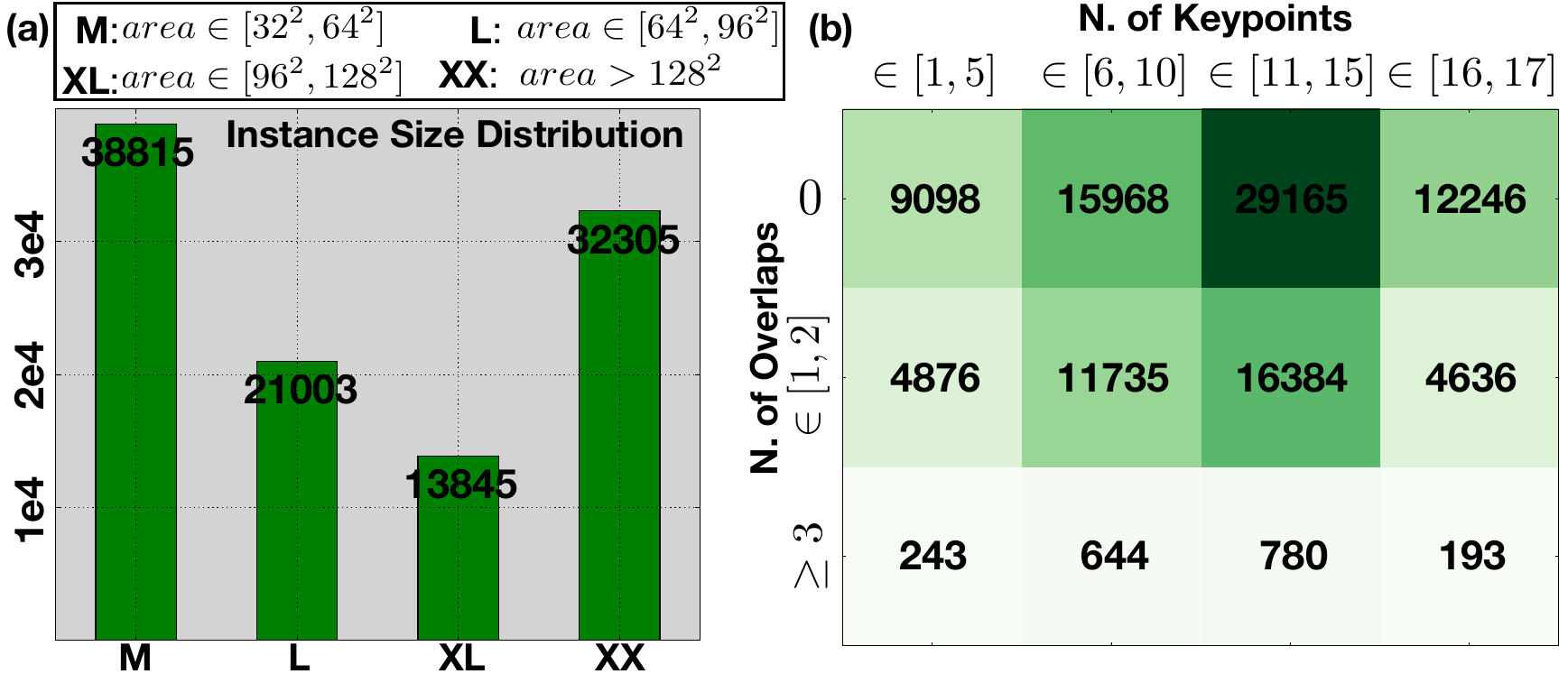}
\caption{ {\small \textbf{Benchmarks of the \textit{COCO} Dataset.} The number of instances in each benchmark of the \textit{COCO} training set based on (a) the size of instances, or (b) the number of overlapping ground-truth annotations with IoU $\ge.1$ and visible keypoints, Fig.~\ref{fig:coco_benchmark}.}}
\label{fig:coco_benchmark_stats}
\vspace{-6mm}
\end{figure}
\vspace{-1mm}
One of the goals of this study is to understand how the layout of people portrayed in images, such as the number of visible keypoints (\textit{occlusion}), the amount of overlap between instances (\textit{crowding}) and size, affects the errors and performance of algorithms. This section is focused on the properties of the data, so we analyze only on method,~\cite{cao2016realtime}. The \textit{COCO} Dataset contains mostly visible instances having little overlap: Fig.~\ref{fig:coco_benchmark_stats} shows that only $1.7\%$ of the annotations have more than two overlaps with an IoU $\ge.1$, and $86.6\%$ have 5 or more visible keypoints. Consequently, we divide the dataset into twelve benchmarks, Fig.~\ref{fig:coco_benchmark}, and study the performance and occurrence of errors in each sepatate one. The PR curves obtained at the evaluation threshold of .75 OKS, after sequentially correcting errors of each type are shown in Fig.~\ref{fig:coco_benchmark_impact}.(a). It appears that the performance of methods listed in Tab.~\ref{tab:eccv_results} is a result of the unbalanced data distribution, and that current algorithms still vastly underperform humans in detecting people and computing their pose, specifically when less than 10 keypoints are visible and overlap is present. Localization errors degrade the performance across all benchmarks, but their impact alone does not explain the shortcomings of current methods. Over $30\%$ of the annotations are missed when the number of visible keypoints is less than 5 (regardless of overlap), and background \textit{FP} and \textit{scoring} errors account for more than $40\%$ of the loss in precision in the benchmarks with high overlap. In Fig.~\ref{fig:coco_benchmark_impact}.(b), we illustrate the frequency of each localization error. \textit{Miss} and \textit{jitter} errors are predominant when there are few keypoints visible, respectively with high and low overlap. \textit{Inversions} are mostly uncorrelated with the amount of overlap, and occur almost always in mostly visible instances. Conversly, \textit{swap} errors depend strongly on the amount of overlap, regardless of the number of visible keypoints. Compared to the overall rates in Fig.~\ref{fig:localization_errors_impact}.(a-cmu) we can see that \textit{inversion} and \textit{jitter} errors are less sensitive to instance overlap and number of keypoints.
\begin{figure*}[t!]
\centering
\includegraphics[width=.9\linewidth]{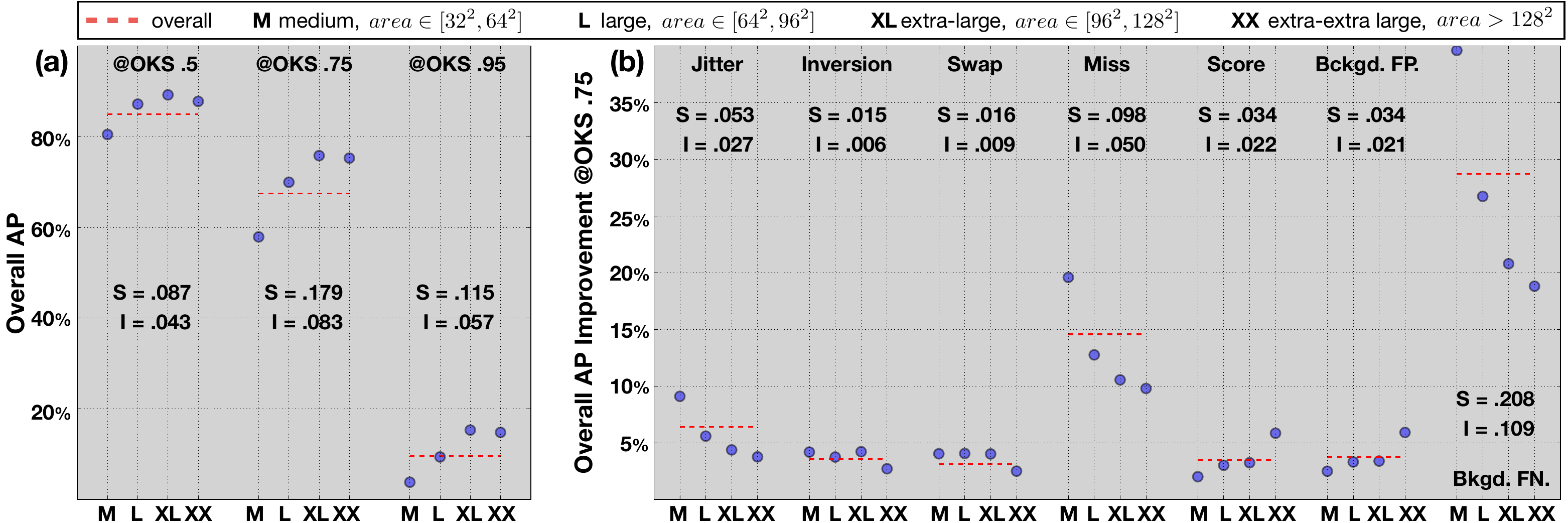}
\caption{ {\small \textbf{Performance and Error Sensitivity to Size.} (a) The overall AP obtained by evaluating~\cite{cao2016realtime} at three OKS evaluation thresholds on the four Size Benchmarks described in Sec.~\ref{sec:coco_benchmark}. (b) The AP improvement at the OKS threshold of .75 obtained after separately correcting each error type on the benchmarks. In both figures, the dashed red line indicates evaluation over all the instance sizes, Sensitivity (S) and Impact (I) are respectively computed as the difference between the maximum and minimum, and the maximum and average, values.}}
\label{fig:size_analysis}
\vspace{-5.5mm}
\end{figure*}
A similar analysis can be done by separating \textit{COCO} into four size groups: \textit{medium}, \textit{large}, \textit{extra-large} and \textit{extra-extra large}, Fig.~\ref{fig:coco_benchmark_stats}.(a). The performance at all OKS evaluation thresholds improves with size, but degrades when instances occupy such a large part of the image that spatial context is lost, Fig.~\ref{fig:size_analysis}.(a). AP is affected by size significantly less than by the amount of overlap and number of visible keypoints. In Fig.~\ref{fig:size_analysis}.(b) we show the AP improvement obtainable by separately correcting each error type in all benchmarks. Errors impact performance less (they occur less often) on larger instances, except for \textit{scoring} and \textit{FP}. Finally, while \textit{FN}, \textit{miss} and \textit{jitter} errors are concentrated on medium instances, all other errors are mostly insensitive to size.

\vspace{-2mm}
\section{Discussion and Recommendations}
\vspace{-1mm}

Multi-instance pose estimation is a challenging visual task where diverse errors have complex causes. Our analysis defines three types of error - \textit{localization}, \textit{scoring}, \textit{background} - and aims to discover and measure their causes, rather than averaging them into a single performance metric. Furthermore, we explore how well a given dataset may be used to probe methods' performance through its statistics of instances' visibility, crowding and size.

The biggest problem for pose estimation is localization errors, present in about 25\% of the predicted keypoints in state of the art methods, Fig.~\ref{fig:localization_errors_impact}.(a). We identify four distinct causes of localization errors,  \textit{Miss}, \textit{Swap}, \textit{Inversion}, and \textit{Jitter}, and study their occurrence in different parts of the body, Fig.~\ref{fig:localization_errors_impact}.(b). The correction of such errors, in particular \textit{Miss}, can bring large improvements in the instance OKS and AP, especially at higher evaluation thresholds, Fig.~\ref{fig:localization_errors_impact}.(c-d).

Another important source of error is noise in the detection's confidence scores. To minimize errors, the scores should be (i) `OKS monotonic increasing' and (ii) calibrated over the whole dataset, Sec.~\ref{subsec:scoring_errors}. The \textit{optimal score} of a given detection corresponds to the maximum OKS value obtainable with any annotation. Replacing a method's scores with the optimal scores yields an average AP improvement of 5\%, Fig.~\ref{fig:scoring_errors_impact}.(a), due to the fact that ground-truth instances match detections that obtain higher OKS, and the overall number of matches is increased, Tab.~\ref{tab:scoring_errors_impact}. A key property of good scoring functions is to separate as much as possible the distribution of confidence scores for detections obtaining high OKS versus low OKS, Fig.~\ref{fig:scoring_errors_impact}.(c).

Characteristics of the portrayed people, such as the amount of overlap with other instances and the number of visible keypoints, substantially affects performance. A comparison between Fig.~\ref{fig:coco_benchmark_impact}.(a) and Tab.~\ref{tab:eccv_results}, shows that average performance strongly depends on the properties of the images, and that state of the art methods still vastly underperform humans when multiple people overlap and significant occlusion is present. Since \textit{COCO} is not rich in such challenging pictures, it remains to be seen whether poor performance is due to the low number of training instances, Fig.~\ref{fig:coco_benchmark_stats}.(b), and a new collection and annotation effort will be needed to investigate this question. The size of instances also affects the quality of the detections, Fig.~\ref{fig:size_analysis}.(a), but is less relevant than occlusion or crowding. This conclusion may be biased by the fact that small instances are not annotated in \textit{COCO} and excluded from our analysis.

In this study we also observe that despite their design differences,~\cite{cao2016realtime,papandreou2017towards} display similar error patterns. Nonetheless, \cite{cao2016realtime} is more sensitive to \textit{swap} errors, as keypoint predictions from the entire image can be erroneously grouped into the same instance, while \cite{papandreou2017towards} is more prone to \textit{misses}, as it only predicts keypoint locations within the detected bounding box. \cite{papandreou2017towards} has more than twice the number of high confidence \textit{FP} errors, compared to \cite{cao2016realtime}. Finally, we observe that \textit{FN} are predominant around the image border for \cite{cao2016realtime}, where grouping keypoints into consistent instances can be harder, and concentrated in the center for \cite{papandreou2017towards}, where there is typically clutter and bounding boxes accuracy is reduced.\\[1ex]
\textbf{Improving Localization:} 3D reasoning along with the estimation of 2D body parts~\cite{taylor2000reconstruction} can improve localization by both incorporating constraints on the anatomical validity of the body part predictions, and learning priors on where to expect visually occluded parts. Two promising directions for improvement are possible: (i) collecting 3D annotations~\cite{bourdev2009poselets} for the humans in \textit{COCO} and learning to directly regress 3D pose end-to-end~\cite{pavlakos2016coarse}; (ii) modeling the manifold of human poses~\cite{akhter2015pose,bogo2016keep,ronchi2016rotation} and learning how to jointly predict the 3D pose of a person along with its 2D skeleton~\cite{tome2017lifting}.\\[1ex]
\textbf{Improving Scoring:} Graphical models~\cite{koller2009probabilistic} can be used to learn a scoring function based on the relative position of body part locations, improving upon~\cite{cao2016realtime,papandreou2017towards} which only use the confidence of the predicted keypoints. Another promising approach is to use the validation set to learn a
regressor for estimating optimal scores (Sec.~\ref{subsec:scoring_errors}) from the confidence maps of the predicted keypoints and from the sub-optimal detection scores generated by the algorithm. Comparing scores of detections in the same image relatively to each other will allow optimizing their order.\\[1ex]
We release our code\footnote{\url{https://goo.gl/9EyDyN}} for future researchers to analyze the strengths and weaknesses of their methods.







{\small
\bibliographystyle{ieee}
\bibliography{egbib}
}

\end{document}